# COMPUTATIONAL MODEL TO GENERATE CASE-INFLECTED FORMS OF MASCULINE NOUNS FOR WORD SEARCH IN SANSKRIT E-TEXT

**[1]Kasmir Raja S.V., [2]V. Rajitha and [3]Meenakshi Lakshmanan**

[1]Dean-Research, SRM University, Chennai, India
[2]Department of Computer Science, Meenakshi College for Women, Chennai, India
Research Scholar, Mother Teresa Women's University, Kodaikanal, India
[3]Head, Department of Computer Science, Meenakshi College for Women, Chennai, India

## ABSTRACT

The problem of word search in Sanskrit is inseparable from complexities that include those caused by euphonic conjunctions and case-inflections. The case-inflectional forms of a noun normally number 24 owing to the fact that in Sanskrit there are eight cases and three numbers-singular, dual and plural. The traditional method of generating these inflectional forms is rather elaborate owing to the fact that there are differences in the forms generated between even very similar words and there are subtle nuances involved. Further, it would be a cumbersome exercise to generate and search for 24 forms of a word during a word search in a large text, using the currently available case-inflectional form generators. This study presents a new approach to generating case-inflectional forms that is simpler to compute. Further, an optimized model that is sufficient for generating only those word forms that are required in a word search and is more than 80% efficient compared to the complete case-inflectional forms generator, is presented in this study for the first time.

## 1. INTRODUCTION

Word search in Sanskrit E-text is a complex problem owing to the phenomena of euphonic conjunctions and case-inflections. These two phenomena in Sanskrit transform words into forms quite different from the original word and hence have to be taken into consideration if a comprehensive word search in E-texts has to be accomplished. This study deals with generating search-related case-inflected forms of words and presents a novel schema and computational algorithm for the same. The authors have already presented (Raja *et al.*, 2014) a solution to the problem with respect to euphonic conjunctions.

A noun or pronoun that is of interest may be encountered variously in a given text depending upon whether it is used in the nominative, accusative, instrumental, dative, genitive, locative or vocative case and whether it is used in the singular, dual or plural. For instance, the basic noun '*div*' (meaning 'heaven') has the form '*dyauḥ*' when it is used in the singular and as the subject of a sentence; when used in the locative case, plural, it has the form '*dyuṣu*' (meaning 'in the heavens'). Thus, a simple search for '*div*' may yield no positive result even though the noun of interest may figure in the text as the subject of a sentence; a search for '*dyauḥ*' would fail to identify all but one of the various

**Corresponding Author:** Meenakshi Lakshmanan, Department of Computer Science, Meenakshi College for Women, Chennai, India

case-inflected forms. Thus, the various case-inflected forms of the term of interest need to be generated and all of them searched for in the target text.

## 2. CASE-INFLECTIONS IN SANSKRIT

There are eight types of case inflections in Sanskrit: Nominative, accusative, instrumental, dative, ablative, possessive, locative and vocative. In each of these, apart from the singular and plural forms of a word that are encountered in most languages, Sanskrit also has a separate dual form. Thus, a word can have up to 24 case-inflected forms. Further, nouns in Sanskrit (and not just the objects they indicate) are categorized into three genders, masculine, feminine and neuter. Case-inflections are defined based on the gender of the word. Different rules for formulating declensions are laid down based on the last letter of the word. For instance, the case-inflections of the word *rāma* (masculine ending in the letter *a*) is done differently compared to the case-inflections of *hari* (masculine ending in the letter *i*). Furthermore, there can be different rules of declension for subspecies of words of a particular gender ending in a specific letter. For example, the word forms generated as case-inflections for the word *dātṛ* (meaning giver) are different from those generated for *bhrātṛ* (meaning brother), though both are masculine words ending in *ṛ*.

Nouns of all the three genders are generally dealt with, in this context, in two categories: Ordinary (*sādhāraṇa-śabda*) and special (*viśeṣa-śabda*). A further categorization within this is that into vowel-ending words and consonant-ending words. The *sādhāraṇa-śabda* of masculine gender of both the vowel-ending and consonant-ending types form the subject matter of this study and the schema and algorithms presented help generate the case-inflected forms for such words.

A sample declension is given in **Table 1**, in both the native *Devanāgarī* script of Sanskrit and the English script, of the *n*-ending masculine word राजन्/*rājan* (meaning king).

## 3. THE PROBLEM COMPLEXITY

A simple search for the word राजन्/*rājan* in a text would clearly not yield the forms राजा/*rājā* or राज्ञे/*rājñe*, which respectively mean 'the king' and 'for/to the king', which one would normally expect to show up in a text search. In fact, out of the 24 inflections, only the vocative singular form matches the original word in this case, while the other 23 have a different spelling altogether; in fact these 23 forms do not even contain the original word as a substring. Thus, unlike in European languages and other Indian languages as well, case-inflections in Sanskrit effect changes in the very form of the word, rather than simply providing extra words to accompany the word or even appending suffixes or prefixes to a basically unchanged word. Thus, in order to carry out a comprehensive search on Sanskrit text, it is imperative that case-inflected forms also be searched for.

The ancient grammarian Pāṇini's seminal work on Sanskrit grammar, the *Aṣṭādhyāyī* (work in eight chapters) (Vāmana and Jayāditya, 1984), devotes 370 aphorisms to case-inflective forms of nouns, as given in the *Siddhānta-kaumudī* (Dīkṣita and Kaumudī, 1962), an authoritative commentary on the *Aṣṭādhyāyī*. The numerous rules laid down in these terse aphorisms are used to build the inflection tables of any given noun. There are thousands of nouns in Sanskrit and they are categorized according to the inflection pattern they follow. For instance, the *n*-ending masculines of the *sādhāraṇa-śabda* class are categorized into seven types, each following different declension rules resulting in seven different types of case-inflection tables.

Another aspect of the problem is the redundancy inherent in the case inflections. It is clear from **Table 1** that there could be duplicates within the 24 inflected forms of the word. These need to be eliminated. It must be stated here that the duplicates occur in different cases and numbers for different types of words.

The challenge in designing an algorithm to generate all the case-inflected forms of a given word that are relevant to the word search scenario, thus involves the correct classification of a word into one of the inflection categories and then the efficient generation of required metamorphosed forms of the word for the purposes of text search.

## 4. BASIS OF THIS STUDY

The book *Śabdamañjarī* by Vidyasagar K. L. V. Sastry and Pandit L. Anantarama Sastri (Vidyasagar *et al.*, 2002), contains the declension tables for Sanskrit nouns belonging to the various categories and is acknowledged widely as a comprehensive consolidation of the relevant Pāṇinian rules. This text has been used as a primary basis for this study, with (Dīkṣita and Kaumudī, 1962;

Vāmana and Jayāditya, 1984) being used to glean further information wherever required.

## 5. DEVELOPMENT OF THE COMPUTATIONAL ALGORITHM

The following control abstraction encapsulates the steps involved in generating the search-related case-inflections.
GenerateInflections (*X, g*)
{// *X* is the given word and *g* its gender
Step 1: Let *x* be the last letter of *X*
Step 2: Find the word category *C* (if any), using *g* and *x*
Step 3: Compute *x'*, the basic transformation of *x* based on *g* and *C*
Step 4: Parse the XML structure to retrieve the formulae for this combination of *x*, *g* and *C*
Step 5: Perform the operations specified in the formulae to generate the inflected forms of *X*}

The input word is taken from the user along with the specification of its gender. As already stated, the computational algorithm consists of two main steps, which are to identify the category of the given word and thereby the operations required to generate the required inflectional forms and to then compute the inflectional forms by performing those operations.

Details of Steps 2-5 are provided in the sections below.

## 6. WORD CATEGORIZATION

Step 2 of the above control abstraction is achieved using a hashing algorithm in order that the process be speeded up considerably. It must be noted here that there is no rule such as one based on letters of the word, etc., that can be used to specify the category of the word. The hash table here consists of words belonging to a specific category organized within the categories of gender and last letter. The hash value is computed based on the gender of the word and its last letter. The table contains word lists and the corresponding category only for those categories of words that have the same gender and last letter but differ in inflected forms. Buckets are used in the hash table for each category to handle collisions, which are inevitable because the input has information only about the gender of the word and its last letter.

## 7. TERMINOLOGY

Before the specification of the formulae developed in this study to compute the inflected forms, it is necessary to introduce some terminology developed exclusively in this study. On studying the inflectional word forms in detail, it was determined that the last part of a word is what changes when an inflectional form is produced, with only the last letter being affected in most cases. Based on this observation, a list of required basic operations on the last letter of words was identified.

For any vowel *x*, the list of operations defined on *x* are listed in **Table 2**.

The consonants of Sanskrit consist of semi-vowels, mutes, sibilants and aspirate. The mutes are given in **Table 3** depicting the columns and rows.

All letters in Columns 1 and 2 are hard consonants and those in Columns 3 and 4 are soft consonants. Column 5 comprises the nasal consonants. 'Softening' a consonant means replacing it by its Column 3 equivalent, i.e., the Column 3 letter in the same row as the given consonant. For instance, softening the letter '*c*' means that it is replaced by the letter '*j*'. Similarly, 'hardening' a consonant means that it is replaced by its Column 1 equivalent. Thus the letter '*g*' when hardened, yields the letter '*k*'. The operation of nasalizing converts the letter into the Column 5 letter lying in the same row.

With this prelude, we now present definitions of functions used in this study. Let *X* be the given search word and let *x* denote its last letter. If *x* is a vowel, let $x_i$ denote the eliding (*lopa*) of *x*, $x_d$ the lengthened form (*dīrgha*) of *x*, $x_g$ the *guṇa* of *x*, $x_u$ the *vṛddhi* of *x*, $x_a$ the *ayāyāvāva* equivalent of *x*, $x_y$ the *yaṇ* equivalent of *x*, as defined in **Table 2**.

If *x* is a consonant, let $x_s$ denote the softened form of *x*, $x_h$ the hardened form of *x*, and $x_n$ the corresponding nasal form of *x*. Let $x_i$, i = 1, 2, … 5 where *x* is a mute, denote the equivalent letter in row *i* of the mutes.

Let $x_c$ denote the result of changing *n* to *ṇ* if required in *X*. This function is based on Pāṇini's aphorism, '*raṣābhyāṁ no ṇaḥ samānapade //8.4.1//*' and two other related aphorisms, which specify conditions under which the letter *n* in a word would be replaced by *ṇ* (Raja *et al.*, 2014).

The operations denoted by the suffixes can be performed in succession and appropriately denoted. For instance, $x_{ga}$ indicates that the *guṇa* operation is first applied to *x* and then the *ayāyāvāva* operation is applied to the resultant. If the first operation is a *lopa*, then the subsequent operation such as in $x_{id}$ implies that the operation denoted by the suffix *d* is applied to the new last letter *x* of *X* got after the eliding operation.

## 8. COMPUTATION OF x'

A detailed study and analysis of the declension tables enumerated as per Pāṇini's rules in (Dīkṣita and Kaumudī,

1962), yielded observations that led to the formation of the formulae for computing the inflectional forms. The formulae were simplified by the introduction of the pre-processing step of computing the value of $x'$ as a derivation from $x$, depending on $g$ and $C$. This step constitutes Step 3 of the control abstraction presented in Section 5 above.

**Table 4** shows how $x'$ is computed from x for all the 35 masculine word categories in the *sādhāraṇa-śabda* list.

Identification of the $x'$ values individually as shown in **Table 4** is a unique approach in the literature and as is clear from the table, enables the option of clubbing of word-endings and categories for the same operation during processing. For instance, the operations for the word categories shown in rows 2,4,5,6 and 7 apply the same operation, $x_g$, to compute $x'$ from $x$.

## 9. DEVELOPMENT OF FORMULAE TO COMPUTE THE INFLECTIONAL FORMS

A list of stems $\delta_i$ that are required to be appended to words in order to produce the inflectional forms, was identified for each category of words, keeping in mind the availability of both the forms $X$ and $X'$ corresponding to $x$ and $x'$ respectively. **Table 5** lists the stems identified.

Though some of the stems in this list can be constructed by appending two or more other stems in the list, such compound stems were not eliminated because they aid in simpler processing. In fact, each letter of the Sanskrit alphabet could have been given a number and considered an atomic stem and compound stems formed from their combinations, but the priority aim of simplifying the final formulae precluded this possibility.

Step 4 of the control abstraction presented in Section 5 above is now explained. A simple XML structure has been developed, which lists the operations required to compute the transformed words for each word category. For example, the following shows the XML structure developed for masculine words ending in '*a*' and those ending in '*i*'. As given in **Table 4**, there are three categories for masculine words ending in '*i*' and the XML structure below groups transformations common to all the three in the higher level of the hierarchy. The transformations are specified as a comma-separated list. The operation '+' in the formulae denote simple string concatenation. A unique feature of this XML structure is that it represents an algorithm by itself apart from acting as a hierarchical organization of data.

```
<Gender G = "masculine">
    <LastLetter L = "a">
```
$x + \delta_1, x' + \delta_2, x' + \delta_4, x + \delta_6, x_d + \delta_7, (x' + \delta_{27})_c, x_d + \delta_{10}, x' + \delta_{12}, x' + \delta_{14}, x' + \delta_{12} + \delta_{16}, x_d + \delta_{17}, x + \delta_{21}, x + \delta_{18}, (x_d + \delta_7 + \delta_{20})_c, x' + \delta_{13}, x' + \delta_{13} + \delta_{25}$
```
    </LastLetter>
    <LastLetter L = "i">
```
$x_d + \delta_7, x + \delta_{10}, x + \delta_{11}, x + \delta_{16}, (x + \delta_{19})_y, (x_d + \delta_7 + \delta_{20})_c, x + \delta_{25}$
```
        <Category C = "1">
```
$x + \delta_1, x_d, x'_a + \delta_3, x + \delta_6, (x + \delta_{28})_c, x'_a + \delta_{13}, x' + \delta_1, x_i + \delta_2$
```
        </Category>
        <Category C = "2">
```
$X' + \delta_3, x'_i, x' + \delta_2, x' + \delta_5, x_g$
```
        </Category>
        <Category C = "3">
```
$x + \delta_1, x_d, x'_a + \delta_3, x + \delta_6, (x + \delta_9)_y, (x + \delta_{13})_y, (x + \delta_{15})_y, (x + \delta_2)_y$
```
        </Category>
    </LastLetter>
    …
</Gender>
```

The forms of the given word $X$ obtained after computing the formulae got on parsing the XML structure, are illustrated in **Table 6**.

As can be seen from the table, there are 16 transformations for words ending in '*a*' and a total of 15, 16 and 16 transformations for Categories 1, 2 and 3 of words ending in '*i*' respectively. This reduction in the number of transformations from 24 as mentioned in Section 2 above has been brought about by eliminating repetitions and by leaving out the original word if it itself appears as an inflectional form. It has been found that over the 35 categories, a reduction in number of transformations can be reduced by about 50-66% by eliminating the redundancies that are inherent in the inflectional forms themselves.

## 10. OPTIMIZATION OF THE COMPUTATIONAL MODEL

Since the requirement is only to find the inflectional forms of words that are needed for a comprehensive word-search and since the '+' operation in the formulae represents string concatenation, formulae of the types $x + \delta_i$ and $x + \delta_i + \delta_j$ contain the original word $X$ as a substring. Hence, for the current application of word search, it is sufficient to consider only those transformations that bring about some change in the word other than appending a string to the word.

**Table 1.** Case inflections of the *n*-ending masculine word *rājan* (meaning king)

| # | Case | Singular | Dual | Plural |
|---|---|---|---|---|
| 1. | Nominative (subject) | राजा *rājā* | राजानौ *rājānau* | राजानः *rājānaḥ* |
| 2. | Accusative (object) | राजानम् *rājānam* | राजानौ *rājānau* | राज्ञः *rājñaḥ* |
| 3. | Instrumental (by, with, through) | राज्ञा *rājñā* | राजभ्याम् *rājabhyām* | राजभिः *rājabhiḥ* |
| 4. | Dative (for, to) | राज्ञे *rājñe* | राजभ्याम् *rājabhyām* | राजभ्यः *rājabhyaḥ* |
| 5. | Ablative (from, than) | राज्ञः *rājñaḥ* | राजभ्याम् *rājabhyām* | राजभ्यः *rājabhyaḥ* |
| 6. | Possessive (belongs to, has/have) | राज्ञः *rājñaḥ* | राज्ञोः *rājñoḥ* | राज्ञाम् *rājñām* |
| 7. | Locative (in, on, at) | राज्ञि/राजनि *rājñi/rājani* | राज्ञोः *rājñoḥ* | राजसु *rājasu* |
| 8. | Vocative (calling out) | (हे) राजन् (*he*)*rājan* | (हे) राजानौ (*he*) *rājānau* | (हे) राजानः (*he*) *rājānaḥ* |

**Table 2.** Operations on vowels

| # | x | dīrgha | guṇa | vṛddhi | ayāyāvāva | yaṇ |
|---|---|---|---|---|---|---|
| 1. | a | ā | a | ā | - | - |
| 2. | ā | ā | - | - | - | - |
| 3. | i | ī | e | ai | - | y |
| 4. | ī | ī | e | ai | - | y |
| 5. | u | ū | o | au | - | v |
| 6. | ū | ū | o | au | - | v |
| 7. | ṛ | ṝ | ar | ār | - | r |
| 8. | ṝ | ṝ | ar | ār | - | r |
| 9. | ḷ | ṝ | al | āl | - | l |
| 10. | e | e | e | ai | ay | - |
| 11. | o | o | o | au | av | - |
| 12. | ai | ai | - | - | āy | - |
| 13. | au | au | - | - | āv | - |

**Table 3.** Table of mute consonants

| # | 1 | 2 | 3 | 4 | 5 |
|---|---|---|---|---|---|
| 1 | k | kh | g | gh | ṅ |
| 2 | c | ch | j | jh | ñ |
| 3 | ṭ | ṭh | ḍ | ḍh | ṇ |
| 4 | t | th | d | dh | n |
| 5 | p | ph | b | bh | m |

This is so because a search for the word '*hari*' in a text, would anyway identify inflectional forms such as '*haribhyām*' and '*haribhyaḥ*' and hence computing such inflectional forms can be discarded. However, such a search would not yield inflectional forms such as '*haraye*', '*harī*', '*hare*' and '*hareḥ*', whereby the corresponding formulae have to be retained.

It may seem from the above examples that a search for the word *X* after performing the single operation $x_i$ would suffice, because anyway *X* is a substring of all the inflectional forms. For example, all the inflectional forms for the word '*hari*' have '*har*' as a substring. However, such trivialization is impossible for many words such as those ending in consonants, as is clear from **Table 4**. The example cited in Section 1 above is also a case in point.

Now the form '*hare*' is represented as *x'* and the form '*hareḥ*' is computed from the formula $x' + \delta_1$. When *X'* with *x'* as its last letter is searched for, all words computed from formulae of the types $x' + \delta_1$ and $x' + \delta_i + \delta_j$ would be found. Similarly, a search for $x_d$ would yield all words computed from formulae $x_d + \delta_i$ and so on.

In the light of the above analysis, the XML structure presented in Section 9 above is reduced to the following, for the same examples of masculine words ending in '*a*' and '*i*'.

<Gender G = "masculine">
  <LastLetter L = "*a*">
    $x' + \delta_2, x_d, x' + \delta_{13}, x' + \delta_{29}$
  </LastLetter>
  <LastLetter L = "*i*">
    $x_d, x_y$
    <Category C = "1">

```
        x'ₐ, x', xᵢ + δ₂
      </Category>
      <Category C = "2">
        x'ₐ, x', xᵢ, x_g
      </Category>
      <Category C = "3">
        x'ₐ, x'
      </Category>
    </LastLetter>
    …
</Gender>
```

As can be seen from the above XML structure, there is a drastic reduction in the number of formulae to be computed. **Table 7** details the number of computations required to generate search-related case-inflections and the percentage of reduction in the number of computations from the XML structure presented in Section 9 above, for all the 35 categories of masculine *sādhāraṇa-śabdas*.

Thus, on average, a reduction of more than 81% of the computations has been achieved through this optimization.

**Table 4.** Computation of *x'* for masculine words

| # | x | Category | x' | Example X | Example X' | Operation to get x' |
|---|---|---|---|---|---|---|
| 1. | a | - | | rāma | rām | $x_i$ |
| 2. | i | 1 | e | hari | hare | $x_g$ |
| 3. | | 2 | y | sakhi | sakhāy | $x_{va}$ |
| 4. | | 3 | e | pati | pate | $x_g$ |
| 5. | u | - | o | guru | guro | $x_g$ |
| 6. | ṛ | 1 | r | pitṛ | pitar | $x_g$ |
| 7. | | 2 | r | nṛ | nar | $x_g$ |
| 8. | | 3 | r | dātṛ | dātār | $x_v$ |
| 9. | ai | - | y | rai | rāy | $x_a$ |
| 10. | o | - | au | go | gau | $x_v$ |
| 11. | au | - | v | glau | glāv | $x_a$ |
| 12. | c | - | k | jalamuc | jalamuk | $x_l$ |
| 13. | j | 1 | k | vaṇij | vaṇik | $x_{h1}$ |
| 14. | | 2 | ṭ | rāj | rāṭ | $x_{ha}$ |
| 15. | t | 1 | t | marut | marud | $x_s$ |
| 16. | | 2 | n | pacat | pacan | $x_n$ |
| 17. | | 3 | n | dhīmat | dhiman | $x_n$ |
| 18. | | 4 | n | mahat | mahan | $x_n$ |
| 19. | d | - | t | suhṛd | suhṛt | $x_h$ |
| 20. | n | 1 | ñ | rājan | rājñ | $x_{ll}$ + "ñ" |
| 21. | | 2 | ā | ātman | ātmā | $x_{id}$ |
| 22. | | 3 | n | śvan | śun | $x_{lll}$ + "un" |
| 23. | | 4 | n | yuvan | yūn | $x_{llld}$ + "n" |
| 24. | | 5 | o | maghavan | maghon | $x_{llll}$ + "on" |
| 25. | | 6 | ā | pathin | panthā | $x_{lll}$ + "nthā" |
| 26. | | 7 | i | karin | kari | $x_l$ |
| 27. | ś | 1 | ṭ | viś | viṭ | $x_l$ + "ṭ" |
| 28. | | 2 | k | tādṛś | tādṛk | $x_l$ + "k" |
| 29. | ṣ | - | ṭ | dviṣ | dviṭ | $x_l$ + "ṭ" |
| 30. | s | 1 | o | vedhas | vedho | $x_{ll}$ + "o" |
| 31. | | 2 | o | śreyas | śreyo | $x_{ll}$ + "o" |
| 32. | | 3 | ṣ | vidvas | viduṣ | $x_{lll}$ + "uṣ" |
| 33. | | 4 | s | pums | pumāṁs | $x_{ll}$ + "māṁs" |
| 34. | | 5 | ṣ | dos | doṣ | $x_l$ + "ṣ" |
| 35. | h | - | ṭ | lih | liṭ | $x_{ll}$ + "ṭ" |

**Table 5.** Index of stems used in creating the inflectional forms

| δ | Stem | δ | Stem | δ | Stem |
|---|---|---|---|---|---|
| 1 | ḥ | 14 | āya | 27 | ena |
| 2 | au | 15 | uḥ | 28 | nā |
| 3 | aḥ | 16 | bhyaḥ | 29 | ai |
| 4 | āḥ | 17 | t | 30 | y |
| 5 | am | 18 | yoḥ | 31 | r |
| 6 | m | 19 | oḥ | 32 | u |
| 7 | n | 20 | ām | 33 | naḥ |
| 8 | a | 21 | sya | 34 | nau |
| 9 | ā | 22 | i | 35 | ān |
| 10 | bhyām | 23 | nām | 36 | nam |
| 11 | bhiḥ | 24 | su | 37 | āṁs |
| 12 | aiḥ | 25 | ṣu | | |
| 13 | e | 26 | āy | | |

**Table 6.** Initial inflectional forms computed for sample words

| # | X | Inflectional forms computed |
|---|---|---|
| 1 | rāma | rāmaḥ, rāmau, rāmāḥ, rāmam, rāmān, rāmeṇa, rāmābhyām, rāmaiḥ, rāmāya, rāmebhyaḥ, rāmāt, rāmasya, rāmayoḥ, rāmāṇām, rāme, rāmeṣu |
| 2 | hari | hariḥ, harī, harayaḥ, harim, harīn, hariṇā, haribhyām, haribhiḥ, haraye, haribhyaḥ, hareḥ, haryoḥ, harīṇām, hariṣu, hare |
| 3 | sakhi | sakhā, sakhāyau, sakhāyaḥ, sakhāyam, sakhīn, sakhyā, sakhibhyām, sakhibhiḥ, sakhye, sakhibhyaḥ, sakhyuḥ, sakhyoḥ, sakhīnām, sakhyau, sakhiṣu, sakhe |
| 4 | pati | patiḥ, patī, patayaḥ, patim patīn, patyā, patibhyām, patibhiḥ, patye, patibhyaḥ, patyuḥ, patyoḥ, patīnām, patyau, patiṣu, pate |

**Table 7.** Optimization statistics for masculine *sādhāraṇa-śabdas*

| # | x | Category | Initial number of computations | Reduced number of computations | Extent of reduction (%) |
|---|---|---|---|---|---|
| 1. | a | - | 16 | 4 | 75.00 |
| 2. | i | 1 | 15 | 5 | 66.67 |
| 3. |  | 2 | 16 | 6 | 62.50 |
| 4. |  | 3 | 16 | 4 | 75.00 |
| 5. | u | - | 15 | 4 | 73.33 |
| 6. | ṛ | 1 | 16 | 3 | 81.25 |
| 7. |  | 2 | 17 | 3 | 82.35 |
| 8. |  | 3 | 16 | 3 | 81.25 |
| 9. | ai | - | 14 | 2 | 85.71 |
| 10. | o | - | 15 | 3 | 80.00 |
| 11. | au | - | 14 | 2 | 85.71 |
| 12. | c | - | 14 | 2 | 85.71 |
| 13. | j | 1 | 14 | 2 | 85.71 |
| 14. |  | 2 | 14 | 2 | 85.71 |
| 15. | t | 1 | 12 | 1 | 91.67 |
| 16. |  | 2 | 13 | 2 | 84.62 |
| 17. |  | 3 | 12 | 3 | 75.00 |
| 18. |  | 4 | 12 | 3 | 75.00 |
| 19. | d | - | 14 | 2 | 85.71 |

Kasmir Raja S.V. *et al*. / Journal of Computer Science 10 (11): 2260-2268, 2014**Table 7.** Continue

| | | | | | |
|---|---|---|---|---|---|
| 20. | *n* | 1 | 15 | 2 | 86.67 |
| 21. | | 2 | 14 | 2 | 85.71 |
| 22. | | 3 | 14 | 3 | 78.57 |
| 23. | | 4 | 14 | 3 | 78.57 |
| 24. | | 5 | 14 | 3 | 78.57 |
| 25. | | 6 | 14 | 3 | 78.57 |
| 26. | | 7 | 16 | 3 | 81.25 |
| 27. | *ś* | 1 | 14 | 2 | 85.71 |
| 28. | | 2 | 13 | 2 | 84.62 |
| 29. | *ṣ* | - | 13 | 2 | 84.62 |
| 30. | *s* | 1 | 14 | 2 | 85.71 |
| 31. | | 2 | 14 | 1 | 92.86 |
| 32. | | 3 | 16 | 3 | 81.25 |
| 33. | | 4 | 16 | 3 | 81.25 |
| 34. | | 5 | 14 | 2 | 85.71 |
| 35. | *h* | - | 14 | 2 | 85.71 |
| Total | | | 504 | 94 | 81.35 |

## 11. CONCLUSION

A new method for computation of case-inflections has been designed for this study, which makes the inflectional form generation more efficient compared to related work in the literature (Huet, 2004a; 2014b; Goyal *et al*., 2012; Huet, 2005; 2009; Goyal and Huet, 2013; Huet, 2003; 2006; 2008; Bhadra *et al*., 2009; Jha and Jha, 2005; Selot *et al*., 2010; Jha *et al*., 2009). The introduction of functions that carry out some basic operations of euphonic conjunctions and the appropriate introduction of a basic transformed form *X'* of the given word as a pre-processing step has greatly enhanced the efficiency of the inflectional form generator.

This efficiency enhancement can be illustrated with a couple of examples. The words *pitṛ* and *dātṛ*, though masculine and ending in the same vowel '*ṛ*', produce different declension tables. The former gives rise to inflectional forms such as *pitarau*, *pitaraḥ* and *pitaram*, while the latter to forms such as *dātārau, dātāraḥ* and *dātāram*. However, before going in for computing the inflectional forms, the algorithm presented in this study generates the *X'* forms, which are, respectively, *pitar* and *dātār* as shown in **Table 4**. Once this is done, there is no difference in the declension formulae for the two words. Similar is the case of formulae with regard to masculine words ending in *c, j, d* and *h* – the formulae are identical once *X'* is appropriately computed as per **Table 4**. Also, masculine words ending in *ś* and *ṣ* are also found to have only one or two dissimilarities with the formulae used to generate the inflectional forms for *c, j, d* and *h*. Hence the algorithm is simplified and there are only a few cases that need to be handled.

Moreover, the optimizing scheme presented above for the sake of generating only those inflectional forms that are required in a search algorithm increases the efficiency by an average of more than 80% as shown. This is unprecedented in the literature.

## 12. ADDITIONAL INFORMATION

### 12.1. Funding Information

No funding agency involved.

### 12.2. Author's Contributions

All authors contributed extensively to the work presented in the paper. All three authors held detailed discussions together in the problem identification and solution conceptualization stages of the work.

**Dr. Kasmir Raja:** Primarily involved in the solution conceptualization and guided the others in the formulation of the control abstraction and in developing the algorithm. He reviewed the work at every stage and provided guidance and shaped the work. He finally thoroughly reviewed and vetted the work.

**Ms. Rajitha V.:** Responsible for formulating the control abstraction and developing the algorithm. She worked out the complete, extensive XML structures, implemented the algorithms and performed rigorous testing.

**Dr. Meenakshi Lakshmanan:** Provided inputs for the Sanskrit language, analysed the efficiency of the algorithm and proposed the optimization of the computational model. She wrote the paper.

### 12.3. Ethics

There are no ethical issues involved in this article.